\documentclass{article}
\pdfoutput=1
\usepackage{ijcai25}

\usepackage{times}
\usepackage{soul}
\usepackage{url}
\usepackage[hidelinks]{hyperref}
\usepackage[utf8]{inputenc}
\usepackage[small]{caption}
\usepackage{graphicx}
\usepackage{amsmath}
\usepackage{amsthm}
\usepackage{booktabs}
\usepackage[switch]{lineno}

\usepackage{pdfpages}
\usepackage[inline, shortlabels]{enumitem}
\usepackage{csquotes}
\usepackage{array}
\usepackage{tikz}
\usepackage{circuitikz}
\usepackage{subcaption}
\usetikzlibrary{arrows, shapes}
\usepackage{amsfonts}
\usepackage{accents}
\usepackage{multirow} %
\usepackage[ruled,vlined,linesnumbered]{algorithm2e}
\SetArgSty{textnormal}
\SetKwProg{Function}{Function}{}{}

\title{A Finite-State Controller Based Offline Solver for Deterministic POMDPs}

\author{
Alex Schutz$^1$
\and
Yang You$^2$\and
Matias Mattamala$^1$\and
Ipek Caliskanelli$^2$\and \\
Bruno Lacerda$^1$\And
Nick Hawes$^1$\\
\affiliations
$^1$University of Oxford\\
$^2$UK Atomic Energy Authority\\
\emails
\{alexschutz, matias, bruno, nickh\}@robots.ox.ac.uk,
\{yang.you, ipek.caliskanelli\}@ukaea.uk
}

\newcommand{\POMDP}{POMDP}
\newcommand{\detPOMDP}{DetPOMDP}
\newcommand{\detMCVI}{DetMCVI}

\usepackage{xcolor}
\definecolor{ponk}{RGB}{229,21,184}

\newcommand{\POMDPSymbol}{\ensuremath{\mathcal{M}}}
\newcommand{\States}{\ensuremath{\mathcal{S}}}
\newcommand{\Actions}{\ensuremath{\mathcal{A}}}
\newcommand{\Transition}{\ensuremath{\mathcal{T}}}
\newcommand{\Observations}{\ensuremath{\mathcal{O}}}
\newcommand{\ObsProbs}{\ensuremath{\mathcal{Z}}}
\newcommand{\Cost}{\ensuremath{c}}
\newcommand{\Goals}{\ensuremath{\mathcal{G}}}
\newcommand{\horizon}{\ensuremath{T}}
\newcommand{\Prob}{\ensuremath{\text{Pr}}}

\newcommand{\belief}{\ensuremath{b}}
\newcommand{\support}{\ensuremath{\text{Supp}}}
\newcommand{\dist}{\ensuremath{\text{dist}}}

\newcommand{\Expect}{\ensuremath{\mathbb{E}}}
\newcommand{\argmax}{\ensuremath{\arg\max}}
\newcommand{\argmin}{\ensuremath{\arg\min}}

\newcommand{\ubar}[1]{\underaccent{\bar}{#1}}

\SetKwRepeat{Do}{do}{while}
\theoremstyle{definition}
\newtheorem{example}{Example}
\newtheorem{definition}{Definition}

\def\thickhline{\noalign{\hrule height.8pt}}

\newcommand*{\x}{\mathsf{x}\mskip1mu}

\begin{document}

\maketitle

\begin{abstract}
    Deterministic partially observable Markov decision processes (\detPOMDP s) often arise in planning problems where the agent is uncertain about its environmental state but can act and observe deterministically.
    In this paper, we propose \detMCVI{}, an adaptation of the Monte Carlo Value Iteration (MCVI) algorithm for \detPOMDP{}s, which builds policies in the form of finite-state controllers~(FSCs).
	\detMCVI{} solves large problems with a high success rate, outperforming existing baselines for \detPOMDP{}s.
    We also verify the performance of the algorithm in a real-world mobile robot forest mapping scenario.
\end{abstract}

\section{Introduction}\label{sec:intro}
\begin{figure}[t]
    \centering\small
    \includegraphics[width=\linewidth]{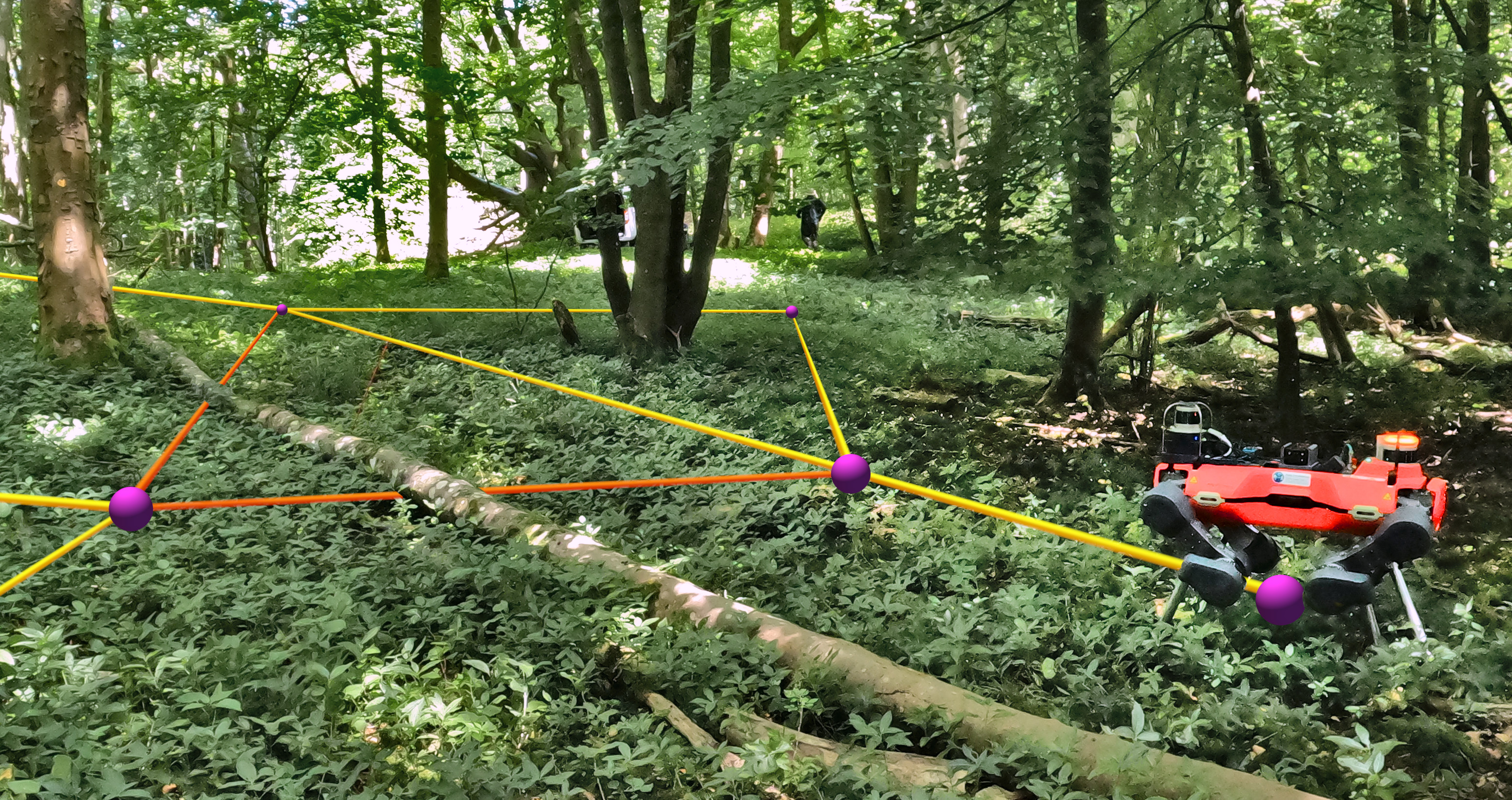}
    \caption{A topological map used for navigation in a forest where possibly obscured terrain leads to uncertain traversability.}
    \label{fig:robot-forest}
\end{figure}

Many planning problems with environmental probabilities are naturally framed as deterministic partially observable Markov decision processes (\detPOMDP{}s), especially where the environment state is not fully known a-priori but can be observed during mission execution.
A common example is robot navigation on a graph where the robot may not know the true traversability of the edges beforehand.
Problems of this type include workplace environments where movement of workers and stock may block routes \cite{nardiLongTermRobotNavigation2020,tsangLAMPLearningMotion2022,lacerdaProbabilisticPlanningFormal2019}, or outdoor environments (see Figure~\ref{fig:robot-forest}) where the traversability of paths is uncertain \cite{huangStochasticPlanningASV2023,deyGaussMeetsCanadian2014}.

\detPOMDP{}s have been under-studied in the literature, with existing approaches relying on the problems being cast as another problem type, such as a general POMDP or an AND-OR graph \cite{bonetDeterministicPOMDPsRevisited2009}.
These approaches have limited applicability on realistic problem sizes.
In many domains, independent uncertainties result in a combinatorial state space.
For example, in the robotic navigation domain, the number of states grows exponentially with the number of uncertain edges.
Such situated AI and robotics problems are typically goal-oriented, as they involve reaching a destination or performing a task.
Furthermore, resource constraints on the robot during navigation often restrict online planning.
Therefore, desirable features of an algorithm for these applications include fast offline synthesis of compact policies with high goal achievement for problems with large state spaces.

In this paper, we introduce \textit{\detMCVI{}}, an offline algorithm designed for goal-oriented \detPOMDP{}s, which achieves state-of-the-art performance on problems with large state spaces.
\detMCVI{} is based on Monte Carlo Value Iteration (MCVI) \cite{baiMonteCarloValue2011}, adapted to goal-oriented settings as per Goal-HSVI \cite{horakGoalHSVIHeuristicSearch2018}, and optimised for deterministic \POMDP{}s.
The algorithm builds policies as finite state controllers (FSCs), which allow for general connectivity in the policy graph and enable compact solutions.%
The FSC structure mitigates the failure cases of tree-based policies when planning is time limited by operational requirements since it allows sub-solutions to be reused, minimising solution incompleteness.
Furthermore, \detMCVI{} scales to domains out of reach of algorithms that require an explicit representation by sampling transitions.
Our implementation is found at \url{http://github.com/ori-goals/DetMCVI}.

The contributions of this paper are:
\begin{enumerate*}[1)]
    \item the introduction of \detMCVI{}, a novel scalable algorithm for solving \detPOMDP{}s;
    \item empirical analysis demonstrating that \detMCVI{} quickly generates compact policies which are more successful than current state-of-the-art baselines;
    \item modelling of a real-world robotics problem involving topological navigation under uncertain environment conditions as a \detPOMDP{}.
\end{enumerate*}

\section{Related Work}\label{sec:related-work}

\subsubsection*{Deterministic POMDPs}
A partially observable Markov decision process (POMDP) is used to model a Markov decision process (MDP) in which the state is not fully observable. 
Interactions with the environment produce \textit{observations}, which inform a \textit{belief} about the current state based on observation probabilities.
A deterministic POMDP is a restriction of a POMDP where actions and observations have deterministic outcomes \cite{bonetDeterministicPOMDPsRevisited2009}.

A POMDP can be modelled as a Belief-MDP, which is an MDP whose states are the possible beliefs of the POMDP.
The number of states in the Belief-MDP of a \detPOMDP{} is upper bounded by $(1 + |\States|)^{|\States|}$ \cite{littmanAlgorithmsSequentialDecisionmaking1996}, making an exact approach generally computationally infeasible.

\citeauthor{bonetDeterministicPOMDPsRevisited2009}~\shortcite{bonetDeterministicPOMDPsRevisited2009} shows that \detPOMDP{}s have a direct relation to AND/OR graphs.
These can be solved offline using search-based heuristic algorithms such as AO$^\star$ \cite{chakrabartiAlgorithmsSearchingExplicit1994}, LAO$^\star$ \cite{hansenLAOHeuristicSearch2001} and RTDP \cite{bartoLearningActUsing1995}, though we do not require the adaptations for cyclic graphs provided by the latter two.
Anytime AO$^{\star}$ \cite{bonetActionSelectionMDPs2021} is a modification which probabilistically searches outside of the best graph, providing better solutions under early termination or with a non-admissible heuristic.
AO$^\star$-based planners have been used for a number of real-world robotics problems \cite{guoRobustCanadianTraveler2019,chungRobotMotionPlanning2011,fergusonPAOPlanningHidden2004}.
These search approaches produce policy trees, and do not leverage similarity in policy features.
In our work we use a more compact policy representation to avoid repeated solving for similar states.

\paragraph{Related Problem Formulations}
A related formulation is the POMDP-lite \cite{chenPOMDPliteRobustRobot2016}, which restricts partial observability to state variables which change deterministically or are constant.
Many of the POMDP-lite domains are examples of \detPOMDP{}s, as \detPOMDP{}s are a restriction of the POMDP-lite. %
Similarly, the Multiple-Environment MDP models problems in which the true environment may be one of many possible MDPs \cite{raskinMultipleEnvironmentMarkovDecision2014}, though no distribution over possible environments is assumed.
A \detPOMDP{} can also be framed as a Bayes-Adaptive MDP (BAMDP) \cite{duffOptimalLearningComputational2002}, where the latent variable encodes the true realisation of the state and the initial distribution is used as the prior.
Conformant planning \cite{bonetConformantPlansPrinciples2010} and Contingent planning \cite{muiseComputingContingentPlans2014,brafmanMultiPathCompilationApproach2021} consider problems with deterministic transitions, partial observability, and an unknown initial state.
They differ from \detPOMDP{}s in that they do not account for a probability distribution over states.

\subsubsection*{Offline POMDP Solutions}
While \detPOMDP{} solution methods have received relatively little attention, POMDP solvers have been heavily researched and optimised, and can be applied directly to \detPOMDP{}s.
The value function of a POMDP can be represented using a finite set of {$\alpha$-vectors} \cite{shaniSurveyPointbasedPOMDP2013}.
Point-based methods generate and optimise a set of {$\alpha$-vectors} to approximate the value function.
Heuristic Search Value Iteration (HSVI) \cite{smithPointBasedPOMDPAlgorithms2005} bounds the values of beliefs in the belief tree to inform heuristics which guide a depth-first search, updating {$\alpha$-vectors} in the backup operation.
SARSOP \cite{kurniawatiSARSOPEfficientPointbased2009} improves on HSVI by focusing on the space of reachable beliefs under optimal policies.
\citeauthor{horakGoalHSVIHeuristicSearch2018}~\shortcite{horakGoalHSVIHeuristicSearch2018} adapt HSVI for use on goal-oriented POMDPs by addressing the lack of convergence guarantees for non-discounted problems, adding a depth bound and preventing re-exploration of action-observation histories.
Point-based approaches require evaluating all {$\alpha$-vectors} in each state, making planning difficult in large state spaces.
Each of these methods requires explicit knowledge of the transition and observation functions to calculate value estimates, thus cannot be used if these functions are unknown or too large to encode.

As an alternative to{ $\alpha$-vectors}, many approaches directly compute an FSC, which represents policies using action nodes and observation edges that lead to the subsequent action node. %
\citeauthor{andriushchenkoInductiveSynthesisFinitestate2022}~\shortcite{andriushchenkoInductiveSynthesisFinitestate2022} search for the best FSC from a set of candidates before expanding the search space in an iterative process, though the approach is limited in scalability as the size of the FSC increases.
Other approaches construct FSCs via non-linear programming \cite{amatoOptimizingFixedsizeStochastic2010}, parametric Markov chains \cite{jungesFinitestateControllersPOMDPs2018}, Anderson acceleration \cite{ermisAndersonAccelerationPartially2021} and belief-integrated FSCs \cite{wrayGeneralizedControllersPOMDP2019}.
However, each of these approaches requires an enumeration of states, which presents scalability barriers for very large state spaces.

Monte Carlo Value Iteration (MCVI) \cite{baiMonteCarloValue2011} iteratively builds an FSC while searching a belief tree using a similar method to SARSOP, calculating value estimates using Monte Carlo simulations.
This approach requires only samples of transitions, and works on continuous-state POMDPs, thus being suitable for large finite state spaces.
In fact, the size of the state space need not be known in advance as states are only accessed from sampled beliefs, in contrast to point-based approaches which evaluate over entire the state space.
These properties make MCVI favourable for specialisation to large \detPOMDP{}s, which we describe in Section~\ref{sec:method}.

For problems with large state-spaces, prior works typically plan online \cite{bonetActionSelectionMDPs2021,eyerichHighqualityPoliciesCanadian2010,silverMonteCarloPlanningLarge2010,chatterjeeMultipleEnvironmentMarkovDecision2020}.
Direct offline application is limited by memory inefficiency.
The QMDP heuristic considers uncertain observations only at the root belief, and assumes a fully observable MDP for child beliefs \cite{littmanLearningPoliciesPartially1995}.
\citeauthor{baiPlanningHowLearn2013}~\shortcite{baiPlanningHowLearn2013} propose a recursive QMDP-based offline policy tree algorithm for solving continuous-state %
robotics problems with deterministic dynamics.
We use QMDP Trees as a baseline for comparison.

\section{Background}\label{sec:background}

\subsection{POMDPs and DetPOMDPs}\label{sec:background:model}
Following~\cite{bonetDeterministicPOMDPsRevisited2009}, we consider the goal-oriented formulation of a \detPOMDP{}, thus also define POMDPs in the goal-oriented setting.

\begin{definition}
A goal-oriented \POMDP{} is a  tuple $\POMDPSymbol = \langle \States, \Actions, \Observations, \belief_0, \Goals, \Transition, \ObsProbs, \Cost \rangle$ where:
\begin{enumerate*}[label=\roman*), nosep]
    \item $\States$ is the set of states;
    \item $\Actions$ is the set of actions;
    \item $\Observations$ is the set of observations;
    \item $\belief_0 \in \Delta(\States)$ is the initial state distribution;
    \item $\Goals \subseteq \States$ is a set of absorbing goal states;
    \item $\Transition: \States \times \Actions \times \States \rightarrow [0, 1]$ is the  transition probability function;
    \item $\ObsProbs: \States \times \Actions \times \Observations \rightarrow [0, 1]$ is the  observation probability function;
    \item $\Cost: \States \times \Actions \rightarrow [0, \infty)$ is the immediate cost of applying action $a$ in state $s$, with $\Cost(s, a) = 0 \iff {s \in \Goals}$. 
\end{enumerate*}

\end{definition}

A belief $b$ for a POMDP is a probability distribution over its state space.
We denote the set of all beliefs over state space $\States$ as~$\Delta(\States)$.
Hereafter, we use $\support(b)$ to indicate the set of states in the support of belief $b$, that is ${\{s \in \States\ |\ b(s) > 0 \}}$.

A solution to a \POMDP{} is a policy $\pi$ which maps an action-observation history ${h_t = (a_0, o_1, a_1, \ldots, a_{t-1}, o_t)}$ to an action $a_t$.
For a policy $\pi$, the value of belief $b$ is given by:
\begin{equation}\label{eq:value-belief}
V^\pi(b) = \Expect_\pi \left[ \sum_{t=0}^\infty \Cost(s_t, a_t)\ |\ b_0 = b \right]\kern-\nulldelimiterspace.
\end{equation}

We seek to minimise the expected cost of the policy given the initial belief $b_0$, i.e., find the policy $\pi$ which minimises $V^\pi(b_0)$.
We denote the value of the belief for the optimal policy as $V^*$.
The action-value function $Q$ relates the expected cost of executing action $a$ in belief $b$ and subsequently following the policy $\pi$:
\begin{equation}\label{eq:q-function}
    Q^\pi(b, a) = \Expect_\pi \left[ \sum_{t=0}^\infty \Cost(s_t, a_t)\ |\ b_0 = b,\ a_0 = a \right]\kern-\nulldelimiterspace.
\end{equation}

The value iteration backup equation constructs a new estimate of the value function from a previous estimate $\tilde{V}$:
\begin{equation}\label{eq:value-iteration}
	\resizebox{.91\linewidth}{!}{$
            \displaystyle
    \tilde{V}'(b) = \min_{a\in\Actions}\left\{ \sum_{s\in\States} \Cost(s, a)b(s) + \sum_{o \in \Observations} \Prob(o | b, a) \tilde{V}(b') \right\}\kern-\nulldelimiterspace.
	$}
\end{equation}
Here,  $b'$ is the subsequent belief which can be calculated using Bayes' rule:

\begin{equation}
\label{eq:belief-update}
\resizebox{.91\linewidth}{!}{$
            \displaystyle
{b'(s') = \tau(b, a, o)(s') = \zeta \ObsProbs(s', a, o) \sum_{s \in \States} \Transition(s, a, s') b(s)},
$}
\end{equation}%
\noindent  where $\zeta$ is a normalisation constant.

In goal-oriented POMDPs, the value function for $\pi$ (Equation~\ref{eq:value-belief}) is finite if and only if $\pi$ reaches a goal state with probability $1$ ($\pi$ is called \textit{proper}).
For convergence, we assume the existence of one such policy \cite{mausamPlanningMarkovDecision2012}.

We represent reachable beliefs in a POMDP as a \emph{belief tree}.

\begin{definition}
    A \textit{belief tree} rooted at $b_0$ is a tree in which nodes represent beliefs ${b \in \Delta(\States)}$, with edges defined by action-observation pairs ${e\in\Actions \times \Observations}$.
    A child node $b'$ is computed from a parent node $b$ via belief update, i.e., if $b'$ is the child of $b$ and the edge from $b$ to $b'$ is $(a,o)$, then $b' = \tau(b,a,o)$, as defined in Equation~\ref{eq:belief-update}. 
\end{definition}

It is not feasible to build a dynamic programming algorithm directly from Equation~\ref{eq:value-iteration}, because the belief space of a POMDP is infinite.
Many approaches have been proposed to approximate solutions to POMDPs, and in this paper we focus on MCVI, which we will present next.
Before doing so, we define the \detPOMDP{} specialisation of a POMDP.

\begin{definition}
A \detPOMDP{} is a POMDP in which the transition and observation functions are deterministic.
We denote the transition function as  $f_{\Transition} : \States \times \Actions \rightarrow \States$, returning the subsequent state after taking action $a$ in state $s$; and the observation function as $f_{\ObsProbs} : \States \times \Actions \rightarrow \Observations$, giving the observation after entering state $s'$ using action $a$.
\end{definition}

The uncertainty in a \detPOMDP{} is only in the initial belief.
Thus, if the state is known exactly at any point in the decision process, the problem is reduced to a deterministic shortest path problem from that point forward.

\begin{example}
\label{exm:ctp}
    The robot navigation problem from the introduction can be posed as a \emph{Canadian Traveller Problem (CTP)} \cite{papadimitriouShortestPathsMap1989}, a topological navigation problem in which some edges are potentially blocked (with a known probability), and the true traversability of the edge can only be observed by the agent at one of the edge's terminal nodes.
    The CTP is a \detPOMDP{} with costs for edge traversal, a goal node, and an initial belief given by the start node and the edge traversability probabilities.
\end{example}

\subsection{Policy Representations}

Policies for POMDPs can be represented in several ways.
In this paper, we are interested in policy trees and FSCs.

\begin{definition}
    A \textit{finite-state controller~(FSC)} is a tuple $F=\langle \mathcal{V}, \eta, \psi \rangle$, where
    \begin{enumerate*}[label=\roman*), nosep]
        \item $\mathcal{V}$ is a finite set of nodes, with start node $v_0$;
        \item $\psi: \mathcal{V} \rightarrow \Actions$ is the action selection function, where $a=\psi(v)$ is the action selected when in node $v$;
        \item $\eta: \mathcal{V} \times \Observations \rightarrow \mathcal{V}$ is the node transition function, where $v'=\eta(v, o)$ is the node transitioned into after observing $o$ in node $v$.
    \end{enumerate*}
\end{definition}

\begin{definition}
    A \textit{policy tree} is an FSC with no cycles and each node $v'$ having at most one node-observation pair $(v,o)$ such that $\eta(v, o) = v'$.
\end{definition}

\subsection{Monte Carlo Value Iteration}\label{sec:background:mcvi}
Monte Carlo Value Iteration (MCVI) \cite{baiMonteCarloValue2011} is an offline method for computing FSCs, designed for continuous-state \POMDP s and therefore also suitable for solving large discrete-state \POMDP s.
The main algorithm for MCVI is a belief tree search, described in Algorithm~\ref{alg:treesearch}.
In this process, a belief tree $T$ is traversed by choosing child beliefs to expand using a guiding heuristic, and then the tree is traversed in reverse order while the bounds at each belief are refined and an FSC $F$ is updated.
The search terminates when the bounds at $b_0$ are within a suitable convergence threshold $\epsilon$.

\begin{algorithm}[t]
	\caption{Belief Tree Search}
	  \label{alg:treesearch}
		\small  \DontPrintSemicolon
\Function{{\upshape \textsc{SearchBeliefTree}($b_0, \epsilon, h$)}}
{
  Initialise an FSC $F$ with an empty set of nodes. \;
  Initialise belief tree $T$ with root $b_0$. \; 
  $\overline{V}(b_0) \leftarrow \infty$,\ 
  $\ubar{V}(b_0) \leftarrow \mathcal{H}(b_0)$. \;
  \While{$\overline{V}(b_0)-\ubar{V}(b_0) > \epsilon$}
  {
    $(b_i)_{i=0}^{k} = ${\upshape \textsc{TraverseBeliefs}($T, b_0, \epsilon, h, F$)} \;
    \For{$i \in k, k-1, \ldots, 0$}
    {
      $F \leftarrow ${\upshape \textsc{Backup}($F, b_i$)}. \label{alg:search:backup}\;
      $\overline{V}(b_i) \leftarrow V^F(b_i)$. \hfill\textit{from~\eqref{eq:value-graph}} \;
      $\ubar{V}(b_i) \leftarrow \ubar{V}'(b_i)$. \hfill\textit{from~\eqref{eq:value-iteration}}\;
    }
  } 
  
  \Return{$F$}
}

\Function{{\upshape \textsc{TraverseBeliefs}($T, b_0, \epsilon, h, F$)}}
{
  $k \leftarrow 0.$ \;
  \Do{$\delta_{b_{k}} > 0$}
  {
    $a^\star \leftarrow \argmin_{a\in\Actions} Q^F(b_k, a)$. \;
    \For{$o\in\Observations$}
      {
        $b^o \leftarrow \tau(b_k, a^\star, o)$. \label{alg:traverse:belief-update}\;
        \If{$b^o \notin T$}
          {
            Add $b^o$ to $T$ with parent $b_k$ and edge $(a^\star, o)$. \;
            $\overline{V}(b^o) \leftarrow \alpha_{F,v_0}(b^o)$,\  
            $\ubar{V}(b^o) \leftarrow \mathcal{H}(b^o)$. \;
          }
        $\delta_{b^o} \leftarrow \overline{V}(b^o)-\ubar{V}(b^o) - \epsilon$. \label{alg:traverse:excess-uncertainty}\;
      }
    $o^\star \leftarrow \argmax_{o\in\Observations} \Prob(o|a^\star, b_k)\delta_{b^o}$. \;
    $b_{k+1} \leftarrow \tau(b_k, a^\star, o^\star)$. \;
    $k \leftarrow k+1$. \;
  }
    \Return{$(b_i)_{i=0}^{k}$}. \;
}

	\end{algorithm}

\begin{figure}[t]
    \centering
    \begin{subfigure}[t]{0.48\columnwidth}
        \centering
            \resizebox{0.9\textwidth}{!}{%
\begin{circuitikz}[scale=0.6]
\node [circle, color=ponk, very thick, draw] (b0) at (13,17.5) {$b_0$} ;
\node [circle, draw] (b2) at (13,11.5) {$b_2$} ;
\node [circle, draw] (b3) at  (16,12.5) {$b_3$} ;
\node [circle, color=ponk, very thick, draw] (b1) at  (10,12.5) {$b_1$} ;
\draw [-o, >=Stealth] (b0) -- (10,15.5) node[circle, inner sep=0, pos=0.5, fill=white]{$a_1$};
\draw [color=ponk, very thick, -o, >=Stealth] (b0) -- (13,14.5)node[circle, inner sep=0, pos=0.5, fill=white, very thick]{$a_2$};
\node [right, font=\small] at (13.5,14.5) {$Q^F(b_0, a_2)$};
\draw [-o, >=Stealth] (b0) -- (16,15.5) node[circle, inner sep=0, pos=0.5, fill=white]{$a_3$};
\draw [->, >=Stealth] (13,14.5) -- (b2) node[circle, inner sep=0, pos=0.5, fill=white]{$o_2$};
\draw [color=ponk, very thick, ->, >=Stealth] (13,14.5) -- (b1)node[circle, inner sep=0, pos=0.5, fill=white, very thick]{$o_1$};
\draw [->, >=Stealth] (13,14.5) -- (b3) node[circle, inner sep=0, pos=0.5, fill=white]{$o_3$};
\node [above, font=\small] at (10,13) {$w_1 (\delta)$};
\draw [dashed] (b1) -- (10,10.5);
\draw [dashed] (b1) -- (9,11);
\draw [dashed] (b1) -- (11,11);
\end{circuitikz}
}%
        \caption{Belief tree}
        \label{fig:tree-graph:tree}
    \end{subfigure}
    \hfill
    \begin{subfigure}[t]{0.48\columnwidth}
        \centering
            \resizebox{0.9\textwidth}{!}{%
\begin{circuitikz}[scale=0.35]
\draw [ rounded corners = 6.0] (11.25,7.5) -- (12.5,6.25) -- (13.75,7.5) -- (12.5,8.75) -- cycle;
\node [circle, inner sep=0.1cm] (a1) at (12.5,7.5) {$a_1$};
\draw [ rounded corners = 6.0] (11.25,13.75) -- (12.5,12.5) -- (13.75,13.75) -- (12.5,15) -- cycle;
\node [circle, inner sep=0.1cm] (a0) at (12.5,13.75) {$a_3$};
\draw [ line width=1.5pt , rounded corners = 6.0] (11.25,18) -- (12.5,16.75) -- (13.75,18) -- (12.5,19.25) -- cycle;
\node [circle, inner sep=0.1cm] (a12) at (12.5,18) {$a_1$};
\draw [ rounded corners = 6.0] (17.5,13.75) -- (18.75,12.5) -- (20,13.75) -- (18.75,15) -- cycle;
\node [circle, inner sep=0.1cm] (a2) at (18.75,13.75) {$a_2$};
\draw [->, >=Stealth] (a12) .. controls (7.5,15.5) and (7.5,12) .. (a1) node[circle, inner sep=0, pos=0.5, fill=white]{$o_3$};
\draw [->, >=Stealth] (a12) -- (a0)node[circle, inner sep=0, pos=0.5, fill=white]{$o_2$};
\draw [->, >=Stealth] (a0) .. controls (11.25,11) and (11.25,10.5) .. (a1) node[circle, inner sep=0, pos=0.5, fill=white]{$o_1$};
\draw [->, >=Stealth] (a2) -- (a0)node[circle, inner sep=0, pos=0.5, fill=white]{$o_3$};
\draw [->, >=Stealth] (a2) -- (a1)node[circle, inner sep=0, pos=0.5, fill=white]{$o_1$};
\draw [->, >=Stealth] (a0) .. controls (13.75,10.75) and (13.75,10.5) .. (a1) node[circle, inner sep=0, pos=0.5, fill=white]{$o_2$};
\draw [->, >=Stealth] (a12) .. controls (16,17.75) and (16.5,17.25) .. (a2) node[circle, inner sep=0, pos=0.5, fill=white]{$o_1$};
\end{circuitikz}
}%
        \caption{FSC}
        \label{fig:tree-graph:graph}
    \end{subfigure}
    \caption{A partial belief tree and FSC built during MCVI iteration.}\label{fig:tree-graph}
\end{figure}
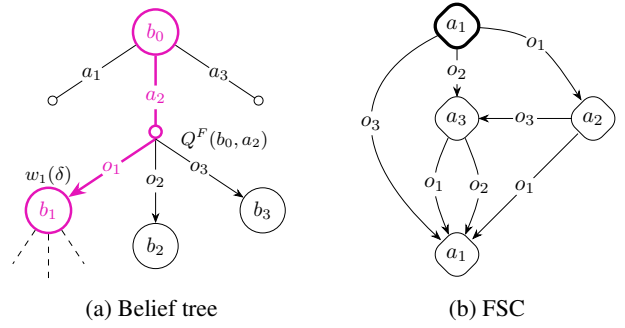

During the tree traversal, actions are chosen to minimise the $Q$ function. 
Observations are chosen which maximise the weighted excess uncertainty of the child belief (line~\ref{alg:traverse:excess-uncertainty}), as per HSVI \cite{smithPointBasedPOMDPAlgorithms2005}.
Child beliefs are generated via particle filtering, with an initial set of $N$ samples (line~\ref{alg:traverse:belief-update}).
The belief tree traversal is illustrated in Figure~\ref{fig:tree-graph:tree}.
Upon reaching a terminal state, or when the excess uncertainty of a belief is negative, the tree is traversed in reverse order, performing a backup operation at each node (line~\ref{alg:search:backup}).

Given an FSC $F$ with starting node $v$, the expected cost of executing $F$ from state $s$ is $\alpha_{F,v}(s)$:
\begin{equation}\label{eq:alpha-vec}
    \alpha_{F,v}(s)=\Expect_{F}\left[ \sum_{t=0}^\infty \Cost(s_t, a_t)\ |\ s_0=s \right]\kern-\nulldelimiterspace.
\end{equation}
We will write $\alpha_{F,v} (b)$ to mean ${\sum_{s\in\support(b)}b(s)\alpha_{F,v}(s)}$.
From \eqref{eq:value-belief}, the value of the belief $b$ under $F$ is:
\begin{equation}\label{eq:value-graph}
    V^F(b) = \min_{v\in \mathcal{V}} \alpha_{F,v}(b).
\end{equation}
In MCVI, $\alpha_{F,v}(s)$ is calculated using Monte Carlo simulations of $F$, given sample-based access to the transition function for subsequent states and observations, with a default policy provided where $F$ is undefined.
An upper bound $\overline{V}(b)$ for $V^*(b)$ is given by $V^\pi(b)$ for any policy $\pi$, so we choose $V^F(b)$ to determine the upper bound using \eqref{eq:value-graph}.
The backup process updates $F$ by adding a new node which improves this upper bound.
Details of the backup process can be found in the appendix.
An example of an FSC built using the MCVI backup process is shown in Figure~\ref{fig:tree-graph:graph}.

Lower bounds are initialised using an admissible heuristic $\mathcal{H}$.
One such heuristic is given by relaxing the POMDP to a fully observable MDP and solving using a suitable MDP method.
The lower bound is updated using the value iteration backup \eqref{eq:value-iteration}, given the lower bounds of the child beliefs.

Note that due to the forward-only construction of the FSC, MCVI cannot generate policies with loops, and is thus unable to support infinite horizon problems without a default policy, which is generally non-trivial to devise.
In general, MCVI is not guaranteed to converge for non-discounted POMDPs \cite{smithProbabilisticPlanningRobotic2007}; we address this in the design of \detMCVI{}.

\section{DetMCVI}\label{sec:method}
There are many inefficiencies when using MCVI to solve \detPOMDP{}s, including re-evaluation of policy rollouts, repeated sampling of deterministic transitions, and storage of beliefs.
We propose \mbox{\detMCVI{}}, an adaptation of MCVI, which solves \detPOMDP{}s efficiently in the goal-oriented setting.

The overall process of \detMCVI{} is similar to Algorithm~\ref{alg:treesearch}, with the main differences in the backup process, presented in Algorithm~\ref{alg:backup}.
Here, value estimates are created for the successor states $s'$ of a belief $b$ by finding the best node in the FSC from which to execute the policy in $s'$ (line \ref{alg:backup:node}).
These value estimates are used to create a new node in the FSC (line \ref{alg:backup:new-fsc}), labelled with the best action, with outgoing edges connecting to the best next node in the FSC for each possible observation.
We will next describe the key differences between the \detMCVI{} backup function and that of MCVI.

\begin{algorithm}[t]
  \caption{\textcolor{black}{Det}MCVI Backup}
  \label{alg:backup}
\small
  \DontPrintSemicolon
\Function{{\upshape \textsc{Backup}($F=\langle \mathcal{V}, \eta, \psi \rangle, b$)}}
{
  For each action $a \in \Actions$, $C_a \leftarrow 0$\textcolor{black}{,\ $\Observations_{b, a} \leftarrow \emptyset$}. \;
  For each action $a \in \Actions$, each observation $o \in \Observations$, and each node $v \in \mathcal{V}$, $V_{a, o, v} \leftarrow 0$.\;
  \For{each action $a \in \Actions$} 
  {
    \For{\textcolor{black}{$s_i \in \support(b)$}\label{alg:backup:state-sample}}
    {
        $s_i' \leftarrow$ \textcolor{black}{$f_{\Transition}(s_i, a)$}. \;
        $o_i \leftarrow$ \textcolor{black}{$f_{\ObsProbs}(s_i', a)$}. \;
        \textcolor{black}{$\Observations_{b, a} \leftarrow \Observations_{b, a} \cup \{o_i \}$.} \;
        $C_a \leftarrow C_a + \textcolor{black}{b(s_i)}\Cost(s_i, a)$.\;
        \For{each node $v \in \mathcal{V}$} 
        {
            Retrieve $\alpha_{F,v}(s'_i)$ from cache or calculate via 1 rollout of the policy $\pi_{F,v}$. \;
            $V_{a, o_i, v} \leftarrow V_{a, o_i, v} + \textcolor{black}{b(s_i)}\alpha_{F,v}(s'_i)$. \label{alg:backup:node}\;
        }
    }
    \For{each observation $o \in \Observations\textcolor{black}{_{b,a}}$} 
    {
          ${V_{a, o} \leftarrow \min_{v\in \mathcal{V}} V_{a, o, v}}$. \;   
		  ${v_{a, o} \leftarrow \argmin_{v\in \mathcal{V}} V_{a, o, v}}$.\;
    }
    $V_a \leftarrow C_a + \gamma \sum_{o\in\Observations\textcolor{black}{_{b,a}}}V_{a, o}$.\;
  }
    $V^{F'} \leftarrow \min_{a \in \Actions} V_a$. \;
    $a^\star \leftarrow \argmin_{a \in \Actions} V_a$.\;
	Create a new FSC $F'=\langle \mathcal{V}', \eta', \psi' \rangle$.
	Set $\psi'(v'_0) = a^\star$ and $\eta'(v'_0, o) = v_{a^\star, o}$. 
    For $k\in 1, \ldots, |\mathcal{V}|$, set $\psi'(v'_{k}) = \psi(v_{k-1})$ and $\eta'(v'_{k}, o) = \eta(v_{k-1}, o)$. \label{alg:backup:new-fsc} \;
  \Return{$F'$}
}

\end{algorithm}

\subsection{Policy Rollouts}
The repeated rollouts required by MCVI when calculating the expected policy value $\alpha_{F,v}$ can be eliminated under deterministic dynamics.
The modifications we make for the policy rollouts are as follows:
\begin{enumerate*}[1)]
	\item we calculate $\alpha_{F,v}(s)$ using a single rollout, which produces the exact value instead of an approximation;
	\item we implement a cache for values of $\alpha_{F,v}(s)$, as the value does not change when $F$ is updated;
	\item instead of calculating the best policy node for each element in the entire set of observations, we restrict the set to only those observations $o$ where $\Prob(o|b, a) > 0$ for a belief $b$ and action $a$.
	This set is calculated during the belief expansion.
	This greatly reduces unnecessary computations, as the size of the full set of observations can be comparatively very large.
\end{enumerate*}

\subsection{Belief Sampling}\label{sec:detmcvi:belief-sample}
The use of Monte Carlo sampling for a deterministic problem is unnecessary, as for any action $a$ each state only has one successor state $s'= f_{\Transition}(s, a)$ and one resultant observation $f_{\ObsProbs}(s', a)$.
Thus, sampling multiple times will return the same result, a property we use for the following changes:
\begin{enumerate*}[1)]
	\item we sample $N$ states and their probabilities from $b_0$ without replacement, instead of sampling $N$ possibly repeated states as in MCVI. 
	\item In line~\ref{alg:backup:state-sample}, we iterate through each state in $\support(b)$ instead of sampling each time, ensuring no duplication.
	\item In a \detPOMDP{}, $|\support(b')| \leq |\support(b)|$ for any successor $b'$ of belief $b$, as all states have one successor and these successors may not be unique.
	Thus, a limit imposed on the maximum size of the initial belief is never exceeded by any descendant beliefs, so $N$ is not imposed in later belief sampling like in MCVI.
\end{enumerate*}

\subsection{Bounds}
As in MCVI, the algorithm performs a search on the belief tree, with upper and lower bounds maintained at each node, until the bounds at the root node converge with a specified tolerance.
Each time a backup is performed, the upper bound $\overline{V}$ is updated according to the value of the belief in the new FSC.
For leaf nodes the lower bound is calculated using an admissible heuristic, for example the full-observability MDP relaxation of the problem.
In a \detPOMDP{} this relaxation reduces the problem to a set of deterministic shortest path problems: 
$\ubar{V}(b) = {\sum_{s\in\support(b)} b(s)\dist(s, \Goals)}$, where $\dist(s, \Goals)$ is the cost of the shortest path to the goal from state $s$.

\subsection{Convergence}
We apply the approach of \citeauthor{horakGoalHSVIHeuristicSearch2018}~\shortcite{horakGoalHSVIHeuristicSearch2018} to guarantee convergence of the algorithm for a goal-oriented \detPOMDP{} under the same conditions used by the authors, namely that the goal is reachable from all states.
	\begin{enumerate*}[1)]
		\item We remove the requirement for a default policy and use uniform random action selection in the rollout calculation of $\alpha_{F,v}$ where $F$ is undefined, with guaranteed termination \cite{chatterjeeOptimalCostAlmostsure2016}. Alternatively, we can remove the requirement for the goal to be reachable from all states so long as rollouts default to a policy which is proper.
		\item We prevent re-exploration of action-observation histories by labelling each node in the belief tree with a binary flag indicating a closed belief.
		This flag is set when all states in the belief are terminal, or when all child beliefs of the node are closed. 
		Closed beliefs are skipped during belief expansion.
\end{enumerate*}
As per \citeauthor{horakGoalHSVIHeuristicSearch2018}~\shortcite{horakGoalHSVIHeuristicSearch2018}, \detMCVI{} attains $\epsilon$-optimality under the conditions when we impose a bound on the search depth $\overline{T}=\frac{\overline{C}}{c_\text{min}}\frac{\overline{C}\eta\epsilon}{(1-\eta)\epsilon}$ for some $\eta < 1$ where $\overline{C}$ is the upper bound on the cost of the uniform policy, and $c_\text{min}$ is the minimum per-step cost.

\section{Synthetic Experiments}\label{sec:results}
\begin{figure}[t]
    \centering
    \begin{subfigure}[t]{0.45\columnwidth}
        \centering
        \includegraphics[width=\textwidth, height=2.4cm, keepaspectratio]{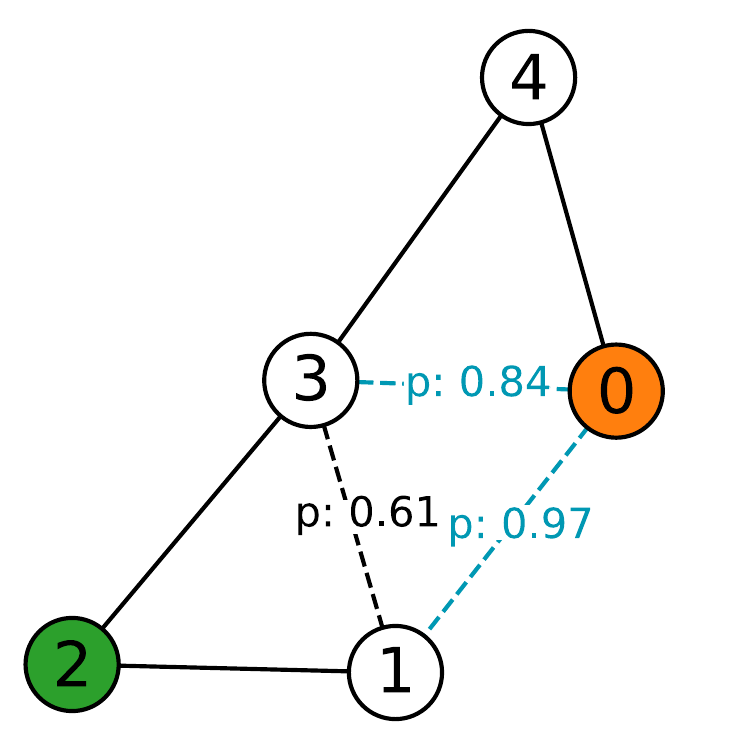}
        \caption{CTP, $n=5$}
        \label{fig:domains:CTP}
    \end{subfigure}\hfill
    \begin{subfigure}[t]{0.45\columnwidth}
        \centering
        \includegraphics[width=\textwidth, height=2.4cm, keepaspectratio]{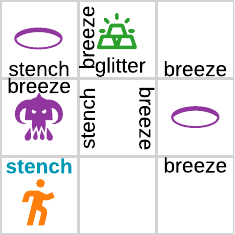}
        \caption{Wumpus, $n=3$}
        \label{fig:domains:wumpus}
    \end{subfigure}\\[0.2em]
    \begin{subfigure}[t]{0.45\columnwidth}
        \centering
        \includegraphics[width=\textwidth, height=2.4cm, keepaspectratio]{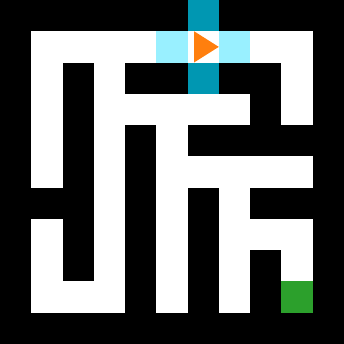}
        \caption{Maze, $n=5$}
        \label{fig:domains:maze}
    \end{subfigure}\hfill
	\begin{subfigure}[t]{0.45\columnwidth}
        \centering
        \includegraphics[width=\textwidth, height=2.4cm, keepaspectratio]{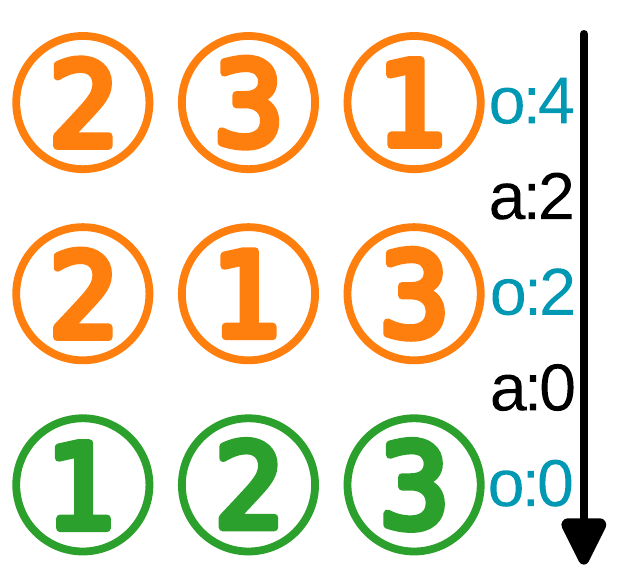}
        \caption{Sort, $n=3$}
        \label{fig:domains:sort}
    \end{subfigure}
    
    \caption{Selected problem instances from each domain}\label{fig:domains}
\end{figure}

We evaluate the performance of \detMCVI{} in different \detPOMDP{} scenarios using the problem domains illustrated in Figure~\ref{fig:domains}.
\begin{enumerate*}[1)]
	\item CTP: from Example~\ref{exm:ctp}. The size of the state space grows exponentially with the number of uncertain edges.
	\item Wumpus World: a goal-oriented modification of that from \citeauthor{russellArtificialIntelligenceModern2021}~\shortcite{russellArtificialIntelligenceModern2021}. Due to the presence of both low- and high-cost goal states, this domain illustrates the importance of optimising the reward function in the process of seeking a goal state.
	\item Maze: this domain has a long horizon but can be solved suboptimally by small FSCs.
	\item Sort: this domain has a short horizon but a solution requires many branches for optimality.
\end{enumerate*}
Full domain descriptions can be found in the appendix.

\begin{table*}[t]
\centering
\resizebox{\textwidth}{!}{
	\begin{scriptsize}
		\bgroup
\setlength\extrarowheight{-1pt}
\setlength{\tabcolsep}{0.5em}
\begin{tabular}{lr|r@{\hspace{0.5\tabcolsep}}lr@{\hspace{0.5\tabcolsep}}lr@{\hspace{0.5\tabcolsep}}l|r@{\hspace{0.5\tabcolsep}}lr@{\hspace{0.5\tabcolsep}}lr@{\hspace{0.5\tabcolsep}}l|r@{\hspace{0.5\tabcolsep}}lr@{\hspace{0.5\tabcolsep}}lr@{\hspace{0.5\tabcolsep}}l|r@{\hspace{0.5\tabcolsep}}lr@{\hspace{0.5\tabcolsep}}l}\thickhline\rule{0pt}{\normalbaselineskip}%
	& & \multicolumn{6}{c|}{\textbf{CTP}} & \multicolumn{6}{c|}{\textbf{Wumpus}} & \multicolumn{6}{c|}{\textbf{Maze}} & \multicolumn{4}{c}{\textbf{Sort}} \\
	& $n$ & \multicolumn{2}{c}{20} & \multicolumn{2}{c}{50} & \multicolumn{2}{c|}{100} & \multicolumn{2}{c}{2} & \multicolumn{2}{c}{3} & \multicolumn{2}{c|}{4} & \multicolumn{2}{c}{10} & \multicolumn{2}{c}{15} & \multicolumn{2}{c|}{20} & \multicolumn{2}{c}{5} & \multicolumn{2}{c}{7} \\
	& $|\States|$ & \multicolumn{2}{c}{$8.60\x 10^{4}$} & \multicolumn{2}{c}{$8.76\x 10^{11}$} & \multicolumn{2}{c|}{$1.19\x 10^{23}$} & \multicolumn{2}{c}{$3.32\x 10^{5}$} & \multicolumn{2}{c}{$7.55\x 10^{7}$} & \multicolumn{2}{c|}{$4.19\x 10^{10}$} & \multicolumn{2}{c}{793} & \multicolumn{2}{c}{1793} & \multicolumn{2}{c|}{3193} & \multicolumn{2}{c}{120} & \multicolumn{2}{c}{5040} \\
   \multicolumn{2}{r|}{$|\support(b_0)|$} & \multicolumn{2}{c}{2790} & \multicolumn{2}{c}{$1.72\x 10^{11} \ddag$} & \multicolumn{2}{c|}{$1.18\x 10^{21} \ddag$} & \multicolumn{2}{c}{72} & \multicolumn{2}{c}{${\geq} 10^{4} \ddag$} & \multicolumn{2}{c|}{${\geq} 7.5\x10^{4} \ddag$} & \multicolumn{2}{c}{792} & \multicolumn{2}{c}{1792} & \multicolumn{2}{c|}{3192} & \multicolumn{2}{c}{119} & \multicolumn{2}{c}{5039} \\
	& $\horizon$ & \multicolumn{2}{c}{40} & \multicolumn{2}{c}{100} & \multicolumn{2}{c|}{200} & \multicolumn{2}{c}{100} & \multicolumn{2}{c}{150} & \multicolumn{2}{c|}{200} & \multicolumn{2}{c}{420} & \multicolumn{2}{c}{930} & \multicolumn{2}{c|}{1640} & \multicolumn{2}{c}{10} & \multicolumn{2}{c}{14} \\\thickhline \rule{0pt}{\normalbaselineskip}%
	MCVI & SR & 60.8 & {\tiny ± 31.4} & \multicolumn{2}{c}{\multirow{4}{*}{†}} & \multicolumn{2}{c|}{\multirow{4}{*}{†}} & \multicolumn{2}{c}{\multirow{4}{*}{*}} & \multicolumn{2}{c}{\multirow{4}{*}{*}} & \multicolumn{2}{c|}{\multirow{4}{*}{*}} & \multicolumn{2}{c}{\multirow{4}{*}{*}} & \multicolumn{2}{c}{\multirow{4}{*}{*}} & \multicolumn{2}{c|}{\multirow{4}{*}{*}} & 20.6 & {\tiny ± 31.0} & 0.8 & {\tiny ± 0.7} \\
	& $\mathcal{R}$ & \multicolumn{2}{c}{-} & & & & & & & & & & & & & & & & & \multicolumn{2}{c}{-} & \multicolumn{2}{c}{-} \\
	& $t_{\text{plan}}$ & 2037.6 & {\tiny ± 1063.2} & & & & & & & & & & & & & & & & & 140.00 & * & 18000 & * \\
	& $|\pi_F|$ & 111 & {\tiny ± 51} & & & & & & & & & & & & & & & & & 126 & {\tiny ± 7} & 721 & {\tiny ± 29} \\\hline \rule{0pt}{\normalbaselineskip}%
   DetMCVI & SR & \textbf{100.0} & \textbf{{\tiny ± 0}} & {\textbf{100.0}} & \textbf{{\tiny ± 0.02}} & {\textbf{99.7}} & \textbf{{\tiny ± 0.5}} & \textbf{100.0} & \textbf{{\tiny ± 0}} & \textbf{93.9} & \textbf{{\tiny ± 4.1}} & \textbf{97.1} & \textbf{{\tiny ± 1.0}} & \textbf{100.0} & \textbf{{\tiny ± 0}} & \textbf{100.0} & \textbf{{\tiny ± 0}} & \textbf{99.5} & \textbf{{\tiny ± 0.8}} & \textbf{100.0} & \textbf{{\tiny ± 0}} & 91.7 & {\tiny ± 0.6} \\
	& $\mathcal{R}$ & 1.184 & {\tiny ± 0.011} & {\textbf{1.18}} & \textbf{{\tiny ± 0.05}} & {1.964} & {\tiny ± 0.070} & 28.4 & {\tiny ± 0.4} & \textbf{118} & \textbf{{\tiny ± 1}} & \textbf{182} & \textbf{{\tiny ± 2}} & 16.1 & {\tiny ± 0.1} & 27.6 & {\tiny ± 0.1} & \textbf{40.2} & \textbf{{\tiny ± 0.1}} & 27.0 & {\tiny ± 0.2} & 52.5 & {\tiny ± 0.3} \\
	& $t_{\text{plan}}$ & \textbf{3.8} & \textbf{{\tiny ± 0.8}} & {\textbf{310}} & \textbf{{\tiny ± 99}} & {\textbf{2594}} & \textbf{{\tiny ± 315}} & 3.78 & {\tiny ± 0.40} & 1200 & * & 18000 & * & 419 & {\tiny ± 128} & 5167 & {\tiny ± 1383} & \textbf{32223} & \textbf{{\tiny ± 5048}} & 2.14 & {\tiny ± 0.02} & 18000 & * \\
	& $|\pi_F|$ & \textbf{11} & \textbf{{\tiny ± 2}} & {\textbf{24}} & \textbf{{\tiny ± 11}} & {\textbf{39}} & \textbf{{\tiny ± 14}} & \textbf{100} & \textbf{{\tiny ± 6}} & \textbf{163} & \textbf{{\tiny ± 61}} & \textbf{636} & \textbf{{\tiny ± 18}} & 493 & {\tiny ± 46} & 1098 & {\tiny ± 113} & \textbf{1647} & \textbf{{\tiny ± 119}} & 164 & {\tiny ± 8} & \textbf{1921} & \textbf{{\tiny ± 70}} \\\hline \rule{0pt}{\normalbaselineskip}%
	AO$^\star$ & SR & \textbf{100.0} & \textbf{{\tiny ± 0}} & {97.3} & {\tiny ± 1.7} & {72.8} & {\tiny ± 18.4} & \textbf{100.0} & \textbf{{\tiny ± 0}} & 7.0 & {\tiny ± 1.2} & 4.8 & {\tiny ± 1.4} & 7.9 & {\tiny ± 2.0} & 4.8 & {\tiny ± 1.2} & 2.1 & {\tiny ± 0.4} & \textbf{100.0} & \textbf{{\tiny ± 0}} & 0.3 & {\tiny ± 0.1} \\
	& $\mathcal{R}$ & \textbf{0.997} & \textbf{{\tiny ± 0.010}} & {\textbf{1.18}} & \textbf{{\tiny ± 0.05}} & {\textbf{0.629}} & \textbf{{\tiny ± 0.023}} & \textbf{22.8} & \textbf{{\tiny ± 0.4}} & \multicolumn{2}{c}{-} & \multicolumn{2}{c|}{-} & \multicolumn{2}{c}{-} & \multicolumn{2}{c}{-} & \multicolumn{2}{c|}{-} & \textbf{24.4} & \textbf{{\tiny ± 0.2}} & \multicolumn{2}{c}{-} \\
	& $t_{\text{plan}}$ & 55.4 & {\tiny ± 72.6} & {901} & {\tiny ± 398} & {18007} & {\tiny ± 8602} & \textbf{0.21} & \textbf{{\tiny ± 0.01}} & 1200 & * & 18000 & * & 1200 & * & 14400 & * & 36000 & * & 59.32 & {\tiny ± 2.23} & 18000 & * \\
	& $|\pi_F|$ & 329 & {\tiny ± 162} & {2768} & {\tiny ± 1394} & {11843} & {\tiny ± 4598} & 163 & {\tiny ± 7} & 127 & {\tiny ± 27} & 97 & {\tiny ± 25} & 3645 & {\tiny ± 387} & 9005 & {\tiny ± 774} & 13434 & {\tiny ± 2213} & 324 & {\tiny ± 0} & 401 & {\tiny ± 38} \\\hline \rule{0pt}{\normalbaselineskip}%
   Anytime & SR & 99.4 & {\tiny ± 1.2} & \multicolumn{2}{c}{\multirow{4}{*}{†}} & \multicolumn{2}{c|}{\multirow{4}{*}{†}} & \textbf{100.0} & \textbf{{\tiny ± 0}} & 1.1 & {\tiny ± 2.2} & 4.3 & {\tiny ± 1.0} & 2.3 & {\tiny ± 0.6} & 1.7 & {\tiny ± 0.4} & 0.9 & {\tiny ± 0.2} & 92.4 & {\tiny ± 13.5} & 0.3 & {\tiny ± 0.3} \\
   AO$^\star$ & $\mathcal{R}$ & \textbf{0.997} & \textbf{{\tiny ± 0.010}} & & & & & \textbf{22.8} & \textbf{{\tiny ± 0.4}} & \multicolumn{2}{c}{-} & \multicolumn{2}{c|}{-} & \multicolumn{2}{c}{-} & \multicolumn{2}{c}{-} & \multicolumn{2}{c|}{-} & 28.7 & {\tiny ± 0.2} & \multicolumn{2}{c}{-} \\
	& $t_{\text{plan}}$ & 33.2 & {\tiny ± 718.2} & & & & & 2.01 & {\tiny ± 0} & 1200 & * & 18000 & * & 1200 & * & 14400 & * & 36000 & * & 96.79 & {\tiny ± 28.25} & 18000 & * \\
	& $|\pi_F|$ & 323 & {\tiny ± 151.8} & & & & & 163 & {\tiny ± 7} & 47 & {\tiny ± 22} & 151 & {\tiny ± 89} & 1969 & {\tiny ± 256} & 5116 & {\tiny ± 397} & 7651 & {\tiny ± 1051} & 350 & {\tiny ± 46} & 224 & {\tiny ± 72} \\\hline \rule{0pt}{\normalbaselineskip}%
   QMDP & SR & \textbf{100.0} & \textbf{{\tiny ± 0.01}} & {97.3} & {\tiny ± 1.7} & {79.6} & {\tiny ± 10.9} & 0 & {\tiny ± 0} & 23.3 & {\tiny ± 0.6} & 15.5 & {\tiny ± 16.4} & 36.6 & {\tiny ± 9.7} & 23.8 & {\tiny ± 2.9} & 17.4 & {\tiny ± 2.0} & \textbf{100.0} & \textbf{{\tiny ± 0}} & \textbf{97.9} & \textbf{{\tiny ± 0.4}} \\
	Trees & $\mathcal{R}$ & \textbf{0.997} & \textbf{{\tiny ± 0.010}} & {\textbf{1.18}} & \textbf{{\tiny ± 0.04}} & {0.817} & {\tiny ± 0.030} & \multicolumn{2}{c}{-} & \multicolumn{2}{c}{-} & \multicolumn{2}{c|}{-} & \multicolumn{2}{c}{-} & \multicolumn{2}{c}{-} & \multicolumn{2}{c|}{-} & 28.9 & {\tiny ± 0.2} & \textbf{41.5} & \textbf{{\tiny ± 0.3}} \\
	& $t_{\text{plan}}$ & 5.9 & {\tiny ± 0.9} & {631} & {\tiny ± 89} & {6135} & {\tiny ± 383} & 0.11 & {\tiny ± 0} & 324 & {\tiny ± 2} & 372 & {\tiny ± 41} & 20 & {\tiny ± 2} & 156 & {\tiny ± 26} & 586 & {\tiny ± 49} & \textbf{0.01} & \textbf{{\tiny ± 0.00}} & \textbf{4} & \textbf{{\tiny ± 0}} \\
	& $|\pi_F|$ & 329 & {\tiny ± 162} & {2768} & {\tiny ± 1394} & {13928} & {\tiny ± 6650} & 401 & {\tiny ± 0} & 2119 & {\tiny ± 0} & 3773 & {\tiny ± 729} & 37992 & {\tiny ± 4938} & 119560 & {\tiny ± 14299} & 260731 & {\tiny ± 15576} & 352 & {\tiny ± 6} & 15819 & {\tiny ± 31} \\\hline \rule{0pt}{\normalbaselineskip}%
   SARSOP & SR & \multicolumn{2}{c}{\multirow{4}{*}{†}} & \multicolumn{2}{c}{\multirow{4}{*}{†}} & \multicolumn{2}{c|}{\multirow{4}{*}{†}} & \multicolumn{2}{c}{\multirow{4}{*}{†}} & \multicolumn{2}{c}{\multirow{4}{*}{†}} & \multicolumn{2}{c|}{\multirow{4}{*}{†}} & \textbf{100.0} & \textbf{{\tiny ± 0}} & \textbf{100.0} & \textbf{{\tiny ± 0}} & \multicolumn{2}{c|}{\multirow{4}{*}{†}} & \textbf{100.0} &  & \multicolumn{2}{c}{\multirow{4}{*}{†}} \\
	& $\mathcal{R}$ & & & & & & & & & & & & & \textbf{5.2} & \textbf{{\tiny ± 0.0}} & \textbf{6.3} & \textbf{{\tiny ± 0.0}} & & & 25.0 & {\tiny ± 0.2} & & \\
	& $t_{\text{plan}}$ & & & & & & & & & & & & & \textbf{8} & \textbf{{\tiny ± 2}} & \textbf{100} & \textbf{{\tiny ± 14}} & & & 0.19 &  & & \\
	& $|\pi_F|$ & & & & & & & & & & & & & \textbf{402} & \textbf{{\tiny ± 28}} & \textbf{949} & \textbf{{\tiny ± 48}} & & & \textbf{107} &  & &\\ \thickhline
	\multicolumn{24}{c}{* Computation time limit reached \ \ $\dagger$ Memory limit reached \ \ $\ddag$ Belief downsampling applied}
   \end{tabular}
\egroup

	\end{scriptsize}
}
\caption{
Evaluation of offline algorithms on large \detPOMDP{} domains. 
SR = success rate (\%), $\mathcal{R}$ = mean regret, $t_{\text{plan}}$ = wall-clock planning time (s), $|\pi_F|$ = number of nodes in policy tree or FSC. 
\detMCVI{} solves the benchmarks with high success rates and small policy sizes.}
\label{tab:results}
\end{table*}

\subsection{Methodology}
In each domain, policies were generated using several baseline solvers.
Specific to \detPOMDP{}s, we evaluate \detMCVI{}, AO$^\star$ \cite{chakrabartiAlgorithmsSearchingExplicit1994}, Anytime AO$^\star$ \cite{bonetActionSelectionMDPs2021}, and QMDP Trees \cite{baiPlanningHowLearn2013}.
We also evaluate general POMDP solvers MCVI \cite{baiMonteCarloValue2011} and SARSOP \cite{kurniawatiSARSOPEfficientPointbased2009}.
For MCVI we use Q-learning for the lower bound heuristic \cite{watkinsLearningDelayedRewards1989} as we assume only sample-based access to the model, while for \detMCVI{}, AO$^\star$ and QMDP Trees we use the bounded-depth Bellman-Ford algorithm to compute the shortest path under full observability without enumerating the entire state space. 
For Anytime AO$^\star$ we use uniform policy rollouts, being faster than the QMDP heuristic but inadmissible.
Implementation details can be found in the appendix.
We impose a domain-dependent horizon $\horizon$ to shorten computation time for practicality.
This means that convergence to an $\epsilon$-optimal policy is not guaranteed, but we find that the policies produced by \detMCVI{} are of sufficient quality long before convergence.

Policies were evaluated at regular intervals using 10000 trials from states randomly sampled from the initial belief.
A trial concluded when a goal state was reached, the horizon was reached, or $\pi(h_t)$ was undefined.
This may occur if an observation is made in a node which does not have provision for that observation, for example when planning does not converge or when downsampling results in states not being planned for.
Planning was terminated after reaching a timeout or memory limitations, or when all trials reached the goal in an evaluation. 
Performance is calculated over sets of 10 problem instances for the CTP and Maze problems, and over three random seeds for Wumpus and Sort.

\paragraph{Belief Downsampling}
For problems where $|\support(b_0)|$ is large, planning can be slow due to processes which operate on all states in the support, such as belief updates and the heuristic calculation.
As described in Section~\ref{sec:detmcvi:belief-sample}, we use a belief for planning which has at most $N$ states in the support.
As the initial belief is only accessed by sampling states, we take $10N$ samples from $b_0$ to create a distribution of relative likelihoods, and choose the first $N$ states from a weighted shuffle and renormalise their probabilities.
We use ${N=10000}$, and we plan with the same sampled initial belief across all baselines.
For the larger CTP and Wumpus problems, $N {<} |\support(b_0)|$, meaning that we do not plan for all states in the support of the initial belief.
The choice of $N$ is analysed in Section~\ref{sec:experiments:belief-downsample}.

\paragraph{Metrics}
We define metrics for a trial beginning in state $s_0$ following policy $\pi$.
The return $R_k$ of a trial up to step $k$ is given by ${\sum_{t=0}^{k-1} \Cost(s_t, \pi(h_t))}$.
For trials where $\pi(h_t)$ is defined for all ${t \in 0, \ldots, \horizon}$, we define the 
full observability \textit{regret} ${\mathcal{R} = R_\horizon - \dist(s_0, \Goals)}$.
We call a trial \textit{successful} if the goal state is reached under $\pi$ at a step $t < \horizon$, and \textit{failed} otherwise.

\subsection{Results}

Results in Table~\ref{tab:results} demonstrate the performance of each algorithm on a selection of different problem instances.
We show the regret for algorithms with a success rate greater than 70\%, and the regret is calculated only for trials from initial states which were successful under all of those algorithms.

The results demonstrate the strong performance of \detMCVI{} in quickly finding very compact solutions which reliably achieve the goal in problems with combinatorial state spaces.
Across the problem domains, \detMCVI{} consistently has a \textbf{high success rate metric}, outperforming the baselines in scalability.
The number of nodes in the policies produced by \detMCVI{} were \textbf{significantly lower} than the policy tree-based baselines, for example by 360 times for CTP with $n=100$, but still competitive in the regret metric.
This difference in policy size is primarily because policy tree-based approaches are not able to reuse existing plans for new branches.
In problems where ${N < |\support(b_0)|}$, this can result in a drop in performance as the policy tree branch is not defined from some states, whereas an FSC policy can still be followed.

\paragraph{Baselines}
SARSOP produced high quality solutions, but could only be applied to the smallest problems due to memory constraints, demonstrating the advantage of a sample-based algorithm.
Due to the lack of a backup process, QMDP Trees prove to be effective in domains where many actions have similar value, so planning for different alternatives does not greatly benefit the solution.
Despite the faster heuristic, Anytime AO$^\star$ did not outperform AO$^\star$ and suffered from memory constraints due to expanding extraneous parts of the belief tree.
MCVI could not produce policies for most of the problems, mainly attributed to the calculation of the heuristic.

\begin{figure}[t]
    \centering
        \includegraphics[width=\linewidth, height=2.9cm, keepaspectratio]{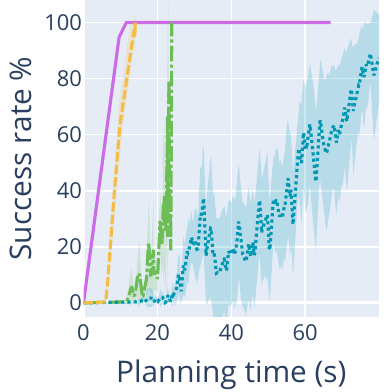}
    \hfill
        \includegraphics[width=\linewidth, height=2.9cm, keepaspectratio]{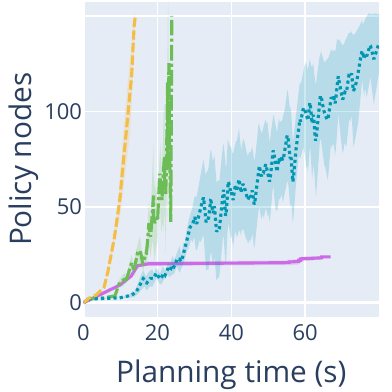}
    \hfill
	\includegraphics[width=\linewidth, height=2.9cm, keepaspectratio]{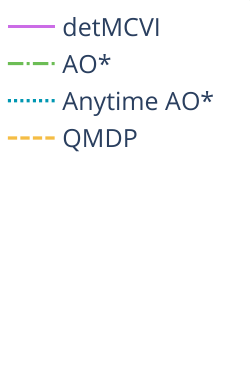}
    \caption{
    Success rate (left) and policy size (right) for different algorithms as evaluated on a CTP problem with $n=25$. 
    }\label{fig:comparison}
\end{figure}

\subsection{Planning with a Budget}
Though offline planning is nominally performed with an infinite time allowance, real-world planning demands time constraints.
It is therefore important to understand the performance of a planning algorithm when terminated early.
Figure~\ref{fig:comparison} shows the evolution of policies with increasing planning time for a CTP problem with $n=25$.
The success rate of the \detMCVI{} policy quickly improves, reaching $100\%$ success rate within $11$ seconds with a policy size of $11$ nodes.
In contrast, AO$^\star$ and QMDP Trees take $24$ and $14$ seconds respectively to achieve a high success rate, and produce policies in excess of $140$ nodes.
Because \detMCVI{} chooses actions according to the upper bound, it is able to quickly find a general solution that reaches the goal from many states, and then improve the cost of the solution for specific states.
Conversely, other \detPOMDP{} planning approaches choose actions using the lower bound, meaning that the goal is not reachable under intermediate policies until planning is complete, even using an anytime method like Anytime AO$^\star$.

\begin{figure}[t]
    \centering
        \includegraphics[width=\linewidth, height=2.9cm, keepaspectratio]{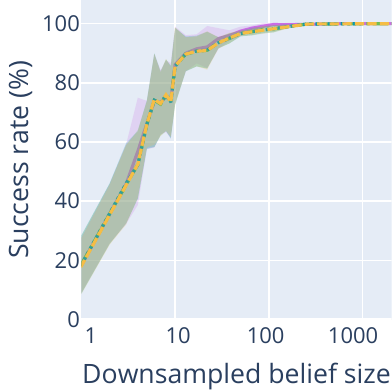}
    \hfill
        \includegraphics[width=\linewidth, height=2.9cm, keepaspectratio]{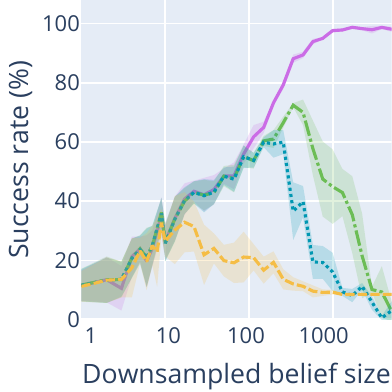}
    \hfill
	\includegraphics[width=\linewidth, height=2.9cm, keepaspectratio]{Figures/downsample_legend.pdf}
    \caption{Success rate of policies generated with downsampled initial beliefs.
	Left: CTP ${n=20}$, ${|\support(b_0)| = 2048}$. Right: Wumpus ${n=3}$, ${|\support(b_0)| \geq 10850}$.
    }
    \label{fig:downsample}
\end{figure}

\subsection{Belief Downsampling}\label{sec:experiments:belief-downsample}
We evaluate the impact of downsampling the initial belief by planning using a range of values for $N$ and evaluating performance over the true initial belief.
Figure~\ref{fig:downsample} demonstrates the success rate of policies for a CTP problem ($n=20$) and a Wumpus problem ($n=3$), noting the logarithmic scale.
In the CTP problem, all algorithms except Anytime AO$^\star$ converged within the time limit, and the planning times increased approximately linearly with $N$.
For the Wumpus problem, \detMCVI{} and AO$^\star$ also began to be affected by the time limit for larger values of $N$.
QMDP Trees performance in Wumpus degrades due to bias toward safe but ineffective actions.
These results show that for these domains, downsampling can offer improvements in computation time, while only affecting solution quality for smaller values of $N$ (${\approx|\support(b_0)|/5}$).

\section{Forest Experiment}\label{sec:field-trial}
The main advantage of \detMCVI{} is fast synthesis of compact policies with high reusability across states. 
This property enables us to use the algorithm on a robot, in a setting where failing to reach the goal is catastrophic, and other algorithms fail to plan sufficiently under the constraints of the real world.
As shown in Figure~\ref{fig:robot-forest}, we evaluate on a robotic navigation problem involving the ANYbotics ANYmal D.
The problem (further described in the appendix) is a modification of the CTP, in which edges can only be observed by attempting to traverse them, rather than being observed from a node.
We use a map generated from operator-guided navigation in a forest, with shortcut edges added for the autonomous phase.

\begin{figure}[t]
    \centering
        \centering
        \includegraphics[width=\textwidth, height=2.7cm, keepaspectratio]{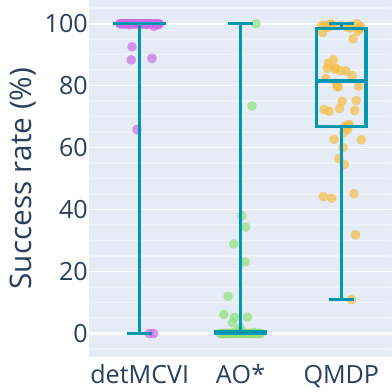}
		\hfill
        \includegraphics[width=\textwidth, height=2.7cm, keepaspectratio]{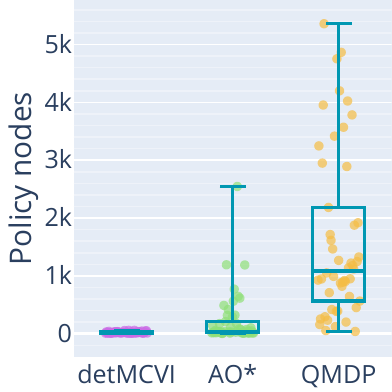}
		\hfill
        \includegraphics[width=\textwidth, height=2.75cm, keepaspectratio]{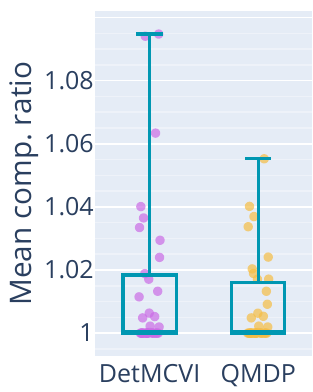}
    \caption{
    Success rate, policy size, and mean competitive ratio over different map realisations from the field data.
    }\label{fig:field_result}
\end{figure}

We evaluate on 50 map instances with randomised edge traversal probabilities and start and goal locations, showing results in Figure~\ref{fig:field_result}.
Anytime AO$^\star$, MCVI and SARSOP failed to return usable policies for the planning budget of 300 seconds.
Across all instances, the average success rates for \detMCVI{}, AO$^\star$, and QMDP Trees respectively were $95\%$, $7\%$, and $78\%$; and the average policy sizes were $24$, $225$ and $1610$.
We use the canonical CTP metric of competitive ratio, defined as the ratio of achieved cost to the best cost under full observability, and we show only the trials which succeeded in both \detMCVI{} and QMDP Trees 
(AO$^\star$ did not have a sufficient success rate to include). 
The average competitive ratios for \detMCVI{} and QMDP Trees were $1.014$ and $1.008$ respectively.
Of the successful trials, \detMCVI{} matches or outperforms the competitive ratio of QMDP Trees on 28 of 35 maps.
The results show that \detMCVI{} nearly always returns a 100\% success rate and has very small policy sizes, with occasionally slightly higher competitive ratios for successful trials.
In the two maps where \detMCVI{} failed to produce a viable policy, all algorithms performed poorly, indicating that the planning budget was not high enough for these instances.

\section{Conclusion}\label{sec:conclusion}
We present \detMCVI{}, an algorithm for solving \detPOMDP{}s in goal-oriented environments.
This algorithm is a simple yet highly effective adaptation of MCVI \cite{baiMonteCarloValue2011} and Goal-HSVI \cite{horakGoalHSVIHeuristicSearch2018}, able to scale to large problems as well as provide convergence guarantees under the same assumptions used by Goal-HSVI.
Empirical evaluations demonstrate that our algorithm obtains competitive performance while scaling to larger problems and producing highly compact policies.
These policies can be several orders of magnitude smaller than state-of-the-art approaches for \detPOMDP{}s, making it beneficial in computationally constrained settings such as on a mobile robot.
Overall, our approach offers a promising method for solving offline, goal-driven problems with deterministic dynamics and uncertain states.
Future work includes augmenting the FSC construction to allow loops, and more efficient heuristic calculations.

\section*{Acknowledgements}
This work received EPSRC funding via the ``From Sensing to Collaboration'' programme grant [EP/V000748/1] and the UKAEA/EPSRC Fusion Grant [EP/W006839/1]. A.S. was supported by a scholarship from the General Sir John Monash Foundation.

\bibliographystyle{named}
\bibliography{references}

\begin{thebibliography}{}

\bibitem[\protect\citeauthoryear{Amato \bgroup \em et al.\egroup
  }{2010}]{amatoOptimizingFixedsizeStochastic2010}
Christopher Amato, Daniel~S. Bernstein, and Shlomo Zilberstein.
\newblock Optimizing fixed-size stochastic controllers for {{POMDPs}} and
  decentralized {{POMDPs}}.
\newblock {\em Autonomous Agents and Multi-Agent Systems}, 21(3):293--320,
  November 2010.

\bibitem[\protect\citeauthoryear{Andriushchenko \bgroup \em et al.\egroup
  }{2022}]{andriushchenkoInductiveSynthesisFinitestate2022}
Roman Andriushchenko, Milan {\v C}e{\v s}ka, Sebastian Junges, and Joost-Pieter
  Katoen.
\newblock Inductive synthesis of finite-state controllers for {{POMDPs}}.
\newblock In {\em Proceedings of the {{Thirty-Eighth Conference}} on
  {{Uncertainty}} in {{Artificial Intelligence}}}, pages 85--95. PMLR, August
  2022.

\bibitem[\protect\citeauthoryear{Bai \bgroup \em et al.\egroup
  }{2011}]{baiMonteCarloValue2011}
Haoyu Bai, David Hsu, Wee~Sun Lee, and Vien~A Ngo.
\newblock Monte {{Carlo}} value iteration for continuous-state {{POMDPs}}.
\newblock In {\em Algorithmic {{Foundations}} of {{Robotics IX}}: {{Selected
  Contributions}} of the {{Ninth International Workshop}} on the {{Algorithmic
  Foundations}} of {{Robotics}}}, pages 175--191. Springer, 2011.

\bibitem[\protect\citeauthoryear{Bai \bgroup \em et al.\egroup
  }{2013}]{baiPlanningHowLearn2013}
Haoyu Bai, David Hsu, and Wee~Sun Lee.
\newblock Planning how to learn.
\newblock In {\em 2013 {{IEEE International Conference}} on {{Robotics}} and
  {{Automation}}}, pages 2853--2859, Karlsruhe, Germany, May 2013. IEEE.

\bibitem[\protect\citeauthoryear{Barto \bgroup \em et al.\egroup
  }{1995}]{bartoLearningActUsing1995}
Andrew~G. Barto, Steven~J. Bradtke, and Satinder~P. Singh.
\newblock Learning to act using real-time dynamic programming.
\newblock {\em Artificial Intelligence}, 72(1-2):81--138, January 1995.

\bibitem[\protect\citeauthoryear{Bonet and
  Geffner}{2021}]{bonetActionSelectionMDPs2021}
Blai Bonet and Hector Geffner.
\newblock Action {{Selection}} for {{MDPs}}: {{Anytime AO}}* {{Versus UCT}}.
\newblock {\em Proceedings of the AAAI Conference on Artificial Intelligence},
  26(1):1749--1755, September 2021.

\bibitem[\protect\citeauthoryear{Bonet}{2009}]{bonetDeterministicPOMDPsRevisited2009}
Blai Bonet.
\newblock Deterministic {{POMDPs Revisited}}.
\newblock In {\em Proceedings of the {{Twenty-Fifth Conference}} on
  {{Uncertainty}} in {{Artificial Intelligence}} ({{UAI}})}, pages 59--66,
  Montreal, Canada, 2009. AUAI Press.

\bibitem[\protect\citeauthoryear{Bonet}{2010}]{bonetConformantPlansPrinciples2010}
Blai Bonet.
\newblock Conformant plans and beyond: {{Principles}} and complexity.
\newblock {\em Artificial Intelligence}, 174(3-4):245--269, March 2010.

\bibitem[\protect\citeauthoryear{Brafman and
  Shani}{2021}]{brafmanMultiPathCompilationApproach2021}
Ronen Brafman and Guy Shani.
\newblock A {{Multi-Path Compilation Approach}} to {{Contingent Planning}}.
\newblock {\em Proceedings of the AAAI Conference on Artificial Intelligence},
  26(1):1868--1874, September 2021.

\bibitem[\protect\citeauthoryear{Chakrabarti}{1994}]{chakrabartiAlgorithmsSearchingExplicit1994}
P.P. Chakrabarti.
\newblock Algorithms for searching explicit {{AND}}/{{OR}} graphs and their
  applications to problem reduction search.
\newblock {\em Artificial Intelligence}, 65(2):329--345, February 1994.

\bibitem[\protect\citeauthoryear{Chatterjee \bgroup \em et al.\egroup
  }{2016}]{chatterjeeOptimalCostAlmostsure2016}
Krishnendu Chatterjee, Martin Chmelik, Raghav Gupta, and Ayush Kanodia.
\newblock Optimal cost almost-sure reachability in {{POMDPs}}.
\newblock {\em Artificial Intelligence}, 234:26--48, 2016.

\bibitem[\protect\citeauthoryear{Chatterjee \bgroup \em et al.\egroup
  }{2020}]{chatterjeeMultipleEnvironmentMarkovDecision2020}
Krishnendu Chatterjee, Martin Chmel{\'i}k, Deep Karkhanis, Petr Novotn{\'y},
  and Am{\'e}lie Royer.
\newblock Multiple-{{Environment Markov Decision Processes}}: {{Efficient
  Analysis}} and {{Applications}}.
\newblock {\em Proceedings of the International Conference on Automated
  Planning and Scheduling}, 30:48--56, June 2020.

\bibitem[\protect\citeauthoryear{Chen \bgroup \em et al.\egroup
  }{2016}]{chenPOMDPliteRobustRobot2016}
Min Chen, Emilio Frazzoli, David Hsu, and Wee~Sun Lee.
\newblock {{POMDP-lite}} for robust robot planning under uncertainty.
\newblock In {\em 2016 {{IEEE International Conference}} on {{Robotics}} and
  {{Automation}} ({{ICRA}})}, pages 5427--5433, Stockholm, Sweden, May 2016.
  IEEE.

\bibitem[\protect\citeauthoryear{Chung and
  Huang}{2011}]{chungRobotMotionPlanning2011}
Shu-Yun Chung and Han-Pang Huang.
\newblock Robot {{Motion Planning}} in {{Dynamic Uncertain Environments}}.
\newblock {\em Advanced Robotics}, 25(6-7):849--870, January 2011.

\bibitem[\protect\citeauthoryear{Dey \bgroup \em et al.\egroup
  }{2014}]{deyGaussMeetsCanadian2014}
Debadeepta Dey, Andrey Kolobov, Rich Caruana, Ece Kamar, Eric Horvitz, and
  Ashish Kapoor.
\newblock Gauss meets {{Canadian}} traveler: Shortest-path problems with
  correlated natural dynamics.
\newblock In {\em {{AAMAS}}'14 {{Proceedings}} of the 2014 {{International
  Conference}} on {{Autonomous Agents}} and {{Multi-agent Systems}}}, pages
  1101--1108, 2014.

\bibitem[\protect\citeauthoryear{Duff}{2002}]{duffOptimalLearningComputational2002}
Michael~O'Gordon Duff.
\newblock {\em Optimal {{Learning}}: {{Computational}} Procedures for
  {{Bayes-adaptive Markov}} Decision Processes}.
\newblock University of Massachusetts Amherst, 2002.

\bibitem[\protect\citeauthoryear{Ermis \bgroup \em et al.\egroup
  }{2021}]{ermisAndersonAccelerationPartially2021}
Melike Ermis, Mingyu Park, and Insoon Yang.
\newblock On {{Anderson Acceleration}} for {{Partially Observable Markov
  Decision Processes}}.
\newblock In {\em 2021 60th {{IEEE Conference}} on {{Decision}} and {{Control}}
  ({{CDC}})}, pages 4478--4485, Austin, TX, USA, December 2021. IEEE.

\bibitem[\protect\citeauthoryear{Eyerich \bgroup \em et al.\egroup
  }{2010}]{eyerichHighqualityPoliciesCanadian2010}
Patrick Eyerich, Thomas Keller, and Malte Helmert.
\newblock High-quality policies for the canadian traveler's problem.
\newblock In {\em Proceedings of the {{AAAI Conference}} on {{Artificial
  Intelligence}}}, volume~24, pages 51--58, 2010.

\bibitem[\protect\citeauthoryear{Ferguson \bgroup \em et al.\egroup
  }{2004}]{fergusonPAOPlanningHidden2004}
D.~Ferguson, A.~Stentz, and S.~Thrun.
\newblock {{PAO}} for planning with hidden state.
\newblock In {\em {{IEEE International Conference}} on {{Robotics}} and
  {{Automation}}, 2004. {{Proceedings}}. {{ICRA}} '04. 2004}, pages 2840--2847
  Vol.3, New Orleans, LA, USA, 2004. IEEE.

\bibitem[\protect\citeauthoryear{Guo and
  Barfoot}{2019}]{guoRobustCanadianTraveler2019}
Hengwei Guo and Timothy~D. Barfoot.
\newblock The {{Robust Canadian Traveler Problem Applied}} to {{Robot
  Routing}}.
\newblock In {\em 2019 {{International Conference}} on {{Robotics}} and
  {{Automation}} ({{ICRA}})}, pages 5523--5529, Montreal, QC, Canada, May 2019.
  IEEE.

\bibitem[\protect\citeauthoryear{Hansen and
  Zilberstein}{2001}]{hansenLAOHeuristicSearch2001}
Eric~A Hansen and Shlomo Zilberstein.
\newblock {{LAO}}*: {{A}} heuristic search algorithm that finds solutions with
  loops.
\newblock {\em Artificial Intelligence}, 129(1-2):35--62, 2001.

\bibitem[\protect\citeauthoryear{Hor{\'a}k \bgroup \em et al.\egroup
  }{2018}]{horakGoalHSVIHeuristicSearch2018}
Karel Hor{\'a}k, Branislav Bosansk{\'y}, and Krishnendu Chatterjee.
\newblock Goal-{{HSVI}}: {{Heuristic Search Value Iteration}} for {{Goal
  POMDPs}}.
\newblock In {\em {{IJCAI}}}, pages 4764--4770, 2018.

\bibitem[\protect\citeauthoryear{Huang \bgroup \em et al.\egroup
  }{2023}]{huangStochasticPlanningASV2023}
Yizhou Huang, Hamza Dugmag, Timothy~D. Barfoot, and Florian Shkurti.
\newblock Stochastic {{Planning}} for {{ASV Navigation Using Satellite
  Images}}.
\newblock In {\em 2023 {{IEEE International Conference}} on {{Robotics}} and
  {{Automation}} ({{ICRA}})}, pages 1055--1061, London, United Kingdom, May
  2023. IEEE.

\bibitem[\protect\citeauthoryear{Junges \bgroup \em et al.\egroup
  }{2018}]{jungesFinitestateControllersPOMDPs2018}
Sebastian Junges, Nils Jansen, Ralf Wimmer, Tim Quatmann, Leonore Winterer,
  Joost-Pieter Katoen, and Bernd Becker.
\newblock Finite-state {{Controllers}} of {{POMDPs}} via {{Parameter
  Synthesis}}.
\newblock {\em Conference on Uncertainty in Artificial Intelligence}, 2018.

\bibitem[\protect\citeauthoryear{Kurniawati \bgroup \em et al.\egroup
  }{2009}]{kurniawatiSARSOPEfficientPointbased2009}
Hanna Kurniawati, David Hsu, and Wee~Sun Lee.
\newblock {{SARSOP}}: {{Efficient Point-Based POMDP Planning}} by
  {{Approximating Optimally Reachable Belief Spaces}}.
\newblock 2009.

\bibitem[\protect\citeauthoryear{Lacerda \bgroup \em et al.\egroup
  }{2019}]{lacerdaProbabilisticPlanningFormal2019}
Bruno Lacerda, Fatma Faruq, David Parker, and Nick Hawes.
\newblock Probabilistic planning with formal performance guarantees for mobile
  service robots.
\newblock {\em The International Journal of Robotics Research},
  38(9):1098--1123, 2019.

\bibitem[\protect\citeauthoryear{Littman \bgroup \em et al.\egroup
  }{1995}]{littmanLearningPoliciesPartially1995}
Michael~L. Littman, Anthony~R. Cassandra, and Leslie~Pack Kaelbling.
\newblock Learning policies for partially observable environments: {{Scaling}}
  up.
\newblock In {\em Machine {{Learning Proceedings}} 1995}, pages 362--370.
  Elsevier, 1995.

\bibitem[\protect\citeauthoryear{Littman}{1996}]{littmanAlgorithmsSequentialDecisionmaking1996}
Michael~Lederman Littman.
\newblock {\em Algorithms for Sequential Decision-Making}.
\newblock Brown University, 1996.

\bibitem[\protect\citeauthoryear{{Mausam} and
  Kolobov}{2012}]{mausamPlanningMarkovDecision2012}
{Mausam} and Andrey Kolobov.
\newblock {\em Planning with {{Markov Decision Processes}}: {{An AI
  Perspective}}}.
\newblock Synthesis {{Lectures}} on {{Artificial Intelligence}} and {{Machine
  Learning}}. Springer International Publishing, Cham, 2012.

\bibitem[\protect\citeauthoryear{Muise \bgroup \em et al.\egroup
  }{2014}]{muiseComputingContingentPlans2014}
Christian Muise, Vaishak Belle, and Sheila McIlraith.
\newblock Computing {{Contingent Plans}} via {{Fully Observable
  Non-Deterministic Planning}}.
\newblock {\em Proceedings of the AAAI Conference on Artificial Intelligence},
  28(1), June 2014.

\bibitem[\protect\citeauthoryear{Nardi and
  Stachniss}{2020}]{nardiLongTermRobotNavigation2020}
Lorenzo Nardi and Cyrill Stachniss.
\newblock Long-{{Term Robot Navigation}} in {{Indoor Environments Estimating
  Patterns}} in {{Traversability Changes}}.
\newblock In {\em 2020 {{IEEE International Conference}} on {{Robotics}} and
  {{Automation}} ({{ICRA}})}, pages 300--306, Paris, France, May 2020. IEEE.

\bibitem[\protect\citeauthoryear{Papadimitriou and
  Yannakakis}{1989}]{papadimitriouShortestPathsMap1989}
Christos~H Papadimitriou and Mihalis Yannakakis.
\newblock Shortest paths without a map.
\newblock In {\em Automata, {{Languages}} and {{Programming}}: 16th
  {{International Colloquium Stresa}}, {{Italy}}, {{July}} 11--15, 1989
  {{Proceedings}} 16}, pages 610--620. Springer, 1989.

\bibitem[\protect\citeauthoryear{Raskin and
  Sankur}{2014}]{raskinMultipleEnvironmentMarkovDecision2014}
Jean-Fran{\c c}ois Raskin and Ocan Sankur.
\newblock Multiple-{{Environment Markov Decision Processes}}, December 2014.

\bibitem[\protect\citeauthoryear{Russell and
  Norvig}{2021}]{russellArtificialIntelligenceModern2021}
S.~Russell and P.~Norvig.
\newblock {\em Artificial {{Intelligence}}: {{A Modern Approach}}, {{Global
  Edition}}}.
\newblock Pearson Education, 2021.

\bibitem[\protect\citeauthoryear{Shani \bgroup \em et al.\egroup
  }{2013}]{shaniSurveyPointbasedPOMDP2013}
Guy Shani, Joelle Pineau, and Robert Kaplow.
\newblock A survey of point-based {{POMDP}} solvers.
\newblock {\em Autonomous Agents and Multi-Agent Systems}, 27(1):1--51, July
  2013.

\bibitem[\protect\citeauthoryear{Silver and
  Veness}{2010}]{silverMonteCarloPlanningLarge2010}
David Silver and Joel Veness.
\newblock Monte-{{Carlo}} planning in large {{POMDPs}}.
\newblock {\em Advances in neural information processing systems}, 23, 2010.

\bibitem[\protect\citeauthoryear{Smith and
  Simmons}{2005}]{smithPointBasedPOMDPAlgorithms2005}
Trey Smith and Reid Simmons.
\newblock Point-based {{POMDP}} algorithms: Improved analysis and
  implementation.
\newblock In {\em Proceedings of the {{Twenty-First Conference}} on
  {{Uncertainty}} in {{Artificial Intelligence}}}, pages 542--549, 2005.

\bibitem[\protect\citeauthoryear{Smith}{2007}]{smithProbabilisticPlanningRobotic2007}
Trey Smith.
\newblock {\em Probabilistic Planning for Robotic Exploration}.
\newblock Carnegie Mellon University, 2007.

\bibitem[\protect\citeauthoryear{Tsang \bgroup \em et al.\egroup
  }{2022}]{tsangLAMPLearningMotion2022}
Florence Tsang, Tristan Walker, Ryan~A. MacDonald, Armin Sadeghi, and
  Stephen~L. Smith.
\newblock {{LAMP}}: {{Learning}} a {{Motion Policy}} to {{Repeatedly Navigate}}
  in an {{Uncertain Environment}}.
\newblock {\em IEEE Transactions on Robotics}, 38(3):1638--1652, June 2022.

\bibitem[\protect\citeauthoryear{Watkins}{1989}]{watkinsLearningDelayedRewards1989}
Christopher John Cornish~Hellaby Watkins.
\newblock Learning from delayed rewards.
\newblock {\em PhD thesis, Cambridge University}, 1989.

\bibitem[\protect\citeauthoryear{Wray and
  Zilberstein}{2019}]{wrayGeneralizedControllersPOMDP2019}
Kyle~Hollins Wray and Shlomo Zilberstein.
\newblock Generalized {{Controllers}} in {{POMDP Decision-Making}}.
\newblock In {\em 2019 {{International Conference}} on {{Robotics}} and
  {{Automation}} ({{ICRA}})}, pages 7166--7172, Montreal, QC, Canada, May 2019.
  IEEE.

\end{thebibliography}

\end{document}